\pdfoutput=1

\documentclass[11pt]{article}

\usepackage[preprint]{acl}

\usepackage{xurl}
\usepackage{times}
\usepackage{latexsym}
\usepackage{kotex}
\usepackage{multicol}
\usepackage{lipsum}
\usepackage{verbatim}
\usepackage{multirow}
\usepackage{xcolor}  
\usepackage{colortbl} 

\usepackage[T1]{fontenc}

\usepackage[utf8]{inputenc}

\usepackage{microtype}

\usepackage{inconsolata}

\usepackage{graphicx}
\usepackage{amsmath}
\usepackage{amsfonts}
\usepackage{algorithm}
\usepackage{algorithmic}

\usepackage{geometry}
\usepackage{longtable}
\usepackage{array}
\usepackage{caption}
\usepackage{booktabs}
\usepackage[most]{tcolorbox}
\usepackage{makecell}  
\usepackage{enumitem}
\usepackage{amssymb}

\title{Retrieval-Augmented Fine-Tuning With Preference Optimization For Visual Program Generation}

\author{
 Deokhyung Kang\textsuperscript{*1}, 
 {\bf Jeonghun Cho\textsuperscript{*1}}, 
 {\bf Yejin Jeon\textsuperscript{1}}, 
 {\bf Sunbin Jang\textsuperscript{3}}, 
\\
 {\bf Minsub Lee\textsuperscript{3}}, 
 {\bf Jawoon Cho\textsuperscript{4}}, 
 {\bf Gary Geunbae Lee\textsuperscript{1,2}}
\\
 \textsuperscript{1}Graduate School of Artificial Intelligence, POSTECH,\\
 \textsuperscript{2}Department of Computer Science and Engineering, POSTECH,\\
 \textsuperscript{3}Hyundai Mobis,
 \textsuperscript{4}T\&I Company
\\
 \texttt{\{deokhk, jeonghuncho, jeonyj0612, gblee\}@postech.ac.kr}\\ \texttt{\{soonbin.Jang, perfect\}@mobis.com}, 
 \texttt{jwcho@tnicompany.com}\\
}

\newenvironment{shk}{%
    \tcblisting{
        enhanced,
        breakable,
        listing only,
        colback=gray!10,
        colframe=black,
        top=2mm,
        bottom=2mm,
        boxrule=0.5pt,
        before skip=10pt,
        after skip=10pt,
        after={\par\vspace{0.5\baselineskip}\noindent},
    }
}
{\endtcblisting}

\begin{document}
\maketitle
\def\thefootnote{*}\footnotetext{Equally contributed}\def\thefootnote{\arabic{footnote}}
\begin{abstract}
Visual programming languages (VPLs) allow users to create programs through graphical interfaces, which results in easier accessibility and their widespread usage in various domains. To further enhance this accessibility, recent research has focused on generating VPL code from user instructions using large language models (LLMs). Specifically, by employing prompting-based methods, these studies have shown promising results. Nevertheless, such approaches can be less effective for industrial VPLs such as Ladder Diagram (LD). LD is a pivotal language used in industrial automation processes and involves extensive domain-specific configurations, which are difficult to capture in a single prompt. In this work, we demonstrate that training-based methods outperform prompting-based methods for LD generation accuracy, even with smaller backbone models. Building on these findings, we propose a two-stage training strategy to further enhance VPL generation. First, we employ retrieval-augmented fine-tuning to leverage the repetitive use of subroutines commonly seen in industrial VPLs. Second, we apply direct preference optimization (DPO) to further guide the model toward accurate outputs, using systematically generated preference pairs through graph editing operations. Extensive experiments on real-world LD data demonstrate that our approach improves program-level accuracy by over 10\% compared to supervised fine-tuning, which highlights its potential to advance industrial automation.
\end{abstract}
\section{Introduction}
Recent advances in large language models (LLMs) have significantly improved their capabilities in code generation. Moreover, LLMs such as GPT-4~\cite{achiam2023gpt}, StarCoder~\cite{li2023starcodersourceyou}, and DeepSeek-Coder~\cite{guo2024deepseekcoderlargelanguagemodel} are able to automate large parts of programming tasks and substantially improve programmers' efficiency. Yet, despite these achievements, most of the previous research has focused on text-based programming languages (TPLs) such as Python or Java, which leaves visual programming languages (VPLs) relatively unexplored.

\begin{figure}
    \centering
    \includegraphics[width=\columnwidth, keepaspectratio]{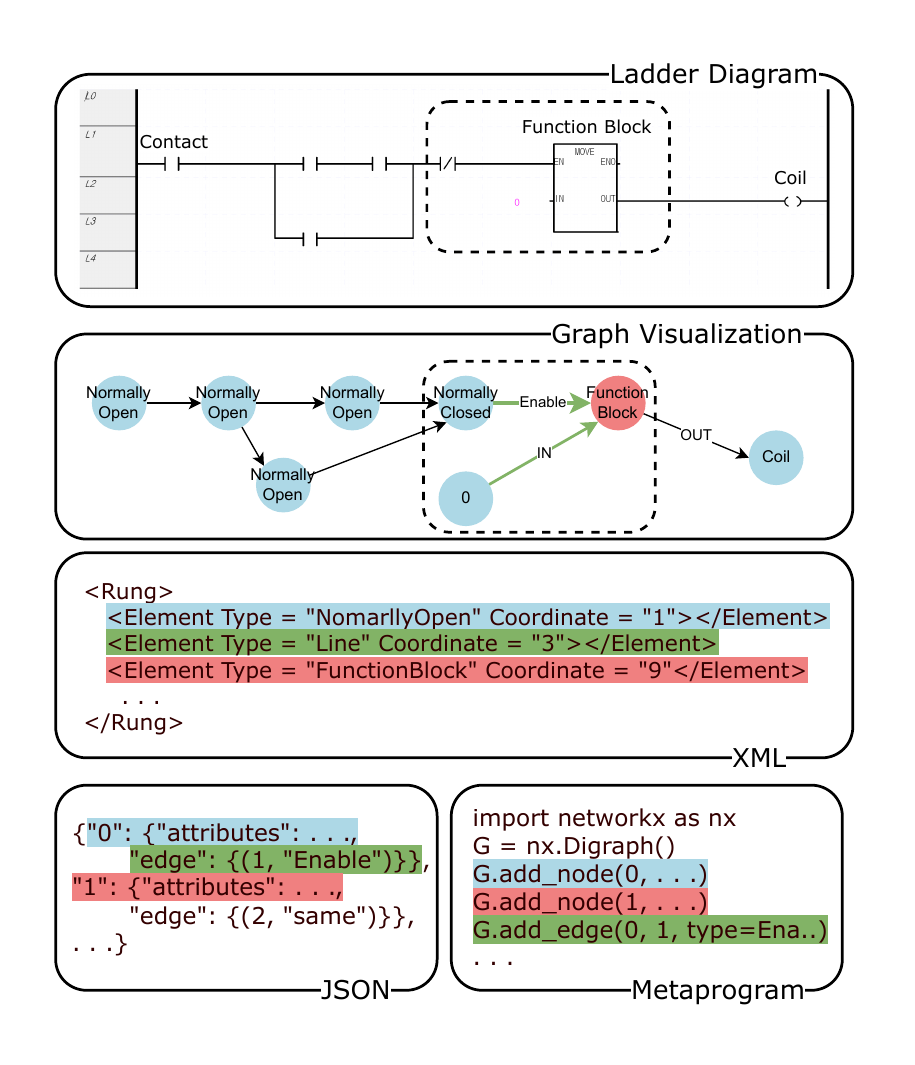}
    \caption{The topmost subfigure shows a single rung of LD and its corresponding visualized graph. The bottom displays XML tags exportable from a Ladder Diagram IDE, along with JSON and Metaprogram representations that capture structural relationships in the graph.}
    \label{fig:format}
\end{figure}

VPLs typically represent programs as node graphs (Figure~\ref{fig:format}), which allow users with limited programming backgrounds to create and modify programs by graphical manipulation~\cite{delozier2023using}. As such, due to their easier accessibility, VPLs have been widely adopted across various domains, from Ladder Diagram (LD) in industrial control systems~\cite{IEC61131-3} to Unreal Engine’s Blueprints in game development~\cite{epicgames_blueprints}. Although VPLs lower the barrier to programming, creating such visual programs from scratch can still be cumbersome. Consequently, recent studies~\cite{zhang2024benchmarking,xue2024comfybenchbenchmarkingllmbasedagents,52868} have explored generating visual programs in text formats (e.g., JSON) from user instructions using prompting-based approaches.

While these studies have shown promising results in VPL generation, the sole reliance on prompting-based methods can be less effective for industrial VPLs like LD. This is because these languages are widely used in industrial automation, which follow domain-specific configurations (e.g., address mapping) that vary drastically by environments~\cite{alphonsus2016review}. Given the vast number of these configurations, it is challenging to include all such information in a single prompt. In contrast, training-based approaches can implicitly learn these configurations during fine-tuning. Therefore, we argue that training-based methods are necessary for this setting.

To validate this, we select LD as a test language and compare supervised fine-tuning (SFT) with retrieval-augmented generation (RAG), where the latter is a commonly used prompting-based method for VPL generation~\cite{zhang2024benchmarking,xue2024comfybenchbenchmarkingllmbasedagents}. Experimental results show that SFT outperforms RAG even with a smaller LLM backbone, which highlights the advantage of training-based approaches for industrial VPL generation.

Motivated by these findings, we propose a two-stage training approach to enhance VPL generation. First, we adopt retrieval-augmented fine-tuning, as VPL often reuses \textit{subroutines} in similar contexts~\cite{Terra_Neves_2021}. During both training and inference, we retrieve similar examples from a corpus and incorporate them into the model input to improve generation quality. Second, we apply offline direct preference optimization (DPO)~\cite{rafailov2023direct} to further guide the model toward accurate outputs.
Since VPLs are represented as node-based graphs, we construct preference pairs by systematically applying graph editing operations to transform ground-truth (preferred) VPL code into unpreferred variants. By learning from these pairs, the model captures the preference relationship between the preferred and unpreferred code, thereby enhancing its fine-grained understanding of VPL's structural properties.

To evaluate our approach, we collect LD data from an actual production environment and conduct extensive evaluations across various text formats and models. Results demonstrate a consistent improvement in generation performance. Notably, our approach improves exact-match accuracy at the program level by more than 10\% compared to SFT, significantly reducing manual inspection efforts and thus accelerating PLC development. Further analyses and ablation studies substantiate the advantages of our methodology. 

Our contributions are as follows:
(1) We present a pioneering study on training-based VPL generation and demonstrate its effectiveness for industrial VPL.
(2) We propose a two-stage training strategy that combines retrieval-based fine-tuning and preference optimization via graph editing, which yields significant performance improvements.
(3) To the best of our knowledge, this is the first work to effectively generate Ladder Diagrams using LLMs, underscoring its potential to enhance industrial automation.

\begin{figure*}[!th]
    \centering
    \includegraphics[width=0.8\textwidth]{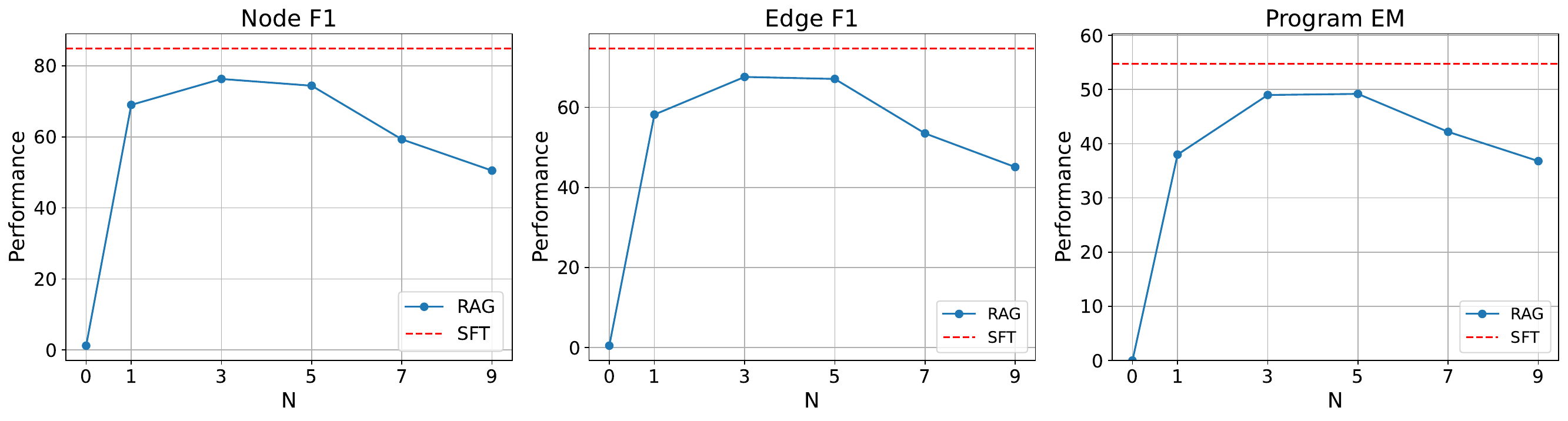} 
    \caption{Performance comparison between SFT and RAG, where RAG uses a larger LLM. $N$ represents the number of retrieved examples in RAG, and SFT’s performance is represented by a red dotted line. We use XML as the text format.}
    \label{fig:rag_sft_performance}
\end{figure*}
\begin{table*}[!ht]
\small
\centering
\resizebox{0.9\textwidth}{!}{%
\begin{tabular}{llccccc}
\toprule
Format                       & Method   & Node F1 $\uparrow$      & Edge F1 $\uparrow$      & Node EM $\uparrow$     & Edge EM $\uparrow$       & Program EM $\uparrow$    \\ \midrule
\multirow{2}{*}{XML}         & RAG & 74.4          & 67.1          & 49.6          & 49.2          & 49.2          \\
                             & SFT      & \textbf{84.8} & \textbf{74.7} & \textbf{55.2} & \textbf{54.8} & \textbf{54.8} \\ \midrule
\multirow{2}{*}{JSON}        & RAG & 72.6          & 65.7          & 51.6          & 50.8          & 50.8          \\
                             & SFT      & \textbf{85.1} & \textbf{74.9} & \textbf{53.6} & \textbf{52.6} & \textbf{52.6} \\ \midrule
\multirow{2}{*}{Metaprogram} & RAG & 80.0            & 71.9          & 52.2          & 51.8          & 51.8          \\
                             & SFT      & \textbf{86.3} & \textbf{75.7} & \textbf{55.2} & \textbf{54.0}   & \textbf{54.0}   \\ \bottomrule
\end{tabular}
}
\caption{Performance across different text formats. The highest values in each format are \textbf{in bold}.}
\label{tab:text_format}
\end{table*}

\section{Is Training-Based Approach Effective?}
In this section, we investigate the effectiveness of the training-based approach in comparison to the prompting-based approach for industrial VPL generation. We introduce Ladder Diagram (LD) as our test language (\S\ref{sec:ld}), then describe text format conversion in \S\ref{sec:conversion} to explore results across different text formats. The evaluation metrics used for comparison are detailed in \S\ref{sec:eval_metric}, and the corresponding results are presented in \S\ref{sec:sft_res}.
\subsection{Ladder Diagram}\label{sec:ld}
Programmable Logic Controller (PLC)~\citep{erickson1996programmable} controls physical devices such as sensors and actuators in industrial automation systems. For example, in a conveyor system, PLC logic enables a robotic arm to detect products via sensors and control its joints to transfer items between conveyors. This logic is typically implemented using Ladder Diagram (LD)~\citep{ladderlogic}, which consists of multiple rungs, each executing a specific task (e.g., operating a motor). Figure~\ref{fig:format} illustrates an LD with a single rung.

Each rung is further composed of contacts, coils, and function blocks and can be represented as a node-based graph, where each visual element is a node. These individual elements serve a distinct role in the logic sequence; contacts control the power flow, coils activate machinery, and function blocks (e.g., timers, counters) provide advanced control functions. They are connected to the PLC and mapped to I/O addresses that vary with the environment and hardware setup. This domain-specific address mapping enables customized control, as each rung manages unique I/O settings tailored to its setup.

\subsection{Text Format Conversion}\label{sec:conversion}
As LLMs cannot generate complex visual elements directly, prior studies~\cite{xue2024comfybenchbenchmarkingllmbasedagents,zhang2024benchmarking} generate VPL in text formats. As each format has unique characteristics, we convert VPL into various text formats to conduct comprehensive experiments. We consider three standard text formats for VPL: (1) XML, which represents visual elements sequentially; (2) JSON, which explicitly captures relationships between elements; (3) Metaprogram (code), which encodes VPL using a code-based syntax. 

Specifically, Figure~\ref{fig:format} illustrates that XML format represents LD as a list of visual elements, including contacts, lines, and function blocks. Each visual element corresponds to an \texttt{<Element>} tag containing its element type, coordinates, and other attributes. Since XML does not explicitly define relationships between elements, we extract these relationships using rule-based methods and convert them into a graph. We iterate through the visual elements in coordinate-based order, adding all elements except lines as nodes to the graph. We implement this graph using NetworkX~\cite{SciPyProceedings_11}, which is a widely used Python library for graph representation. We assign Node IDs starting from 0, increasing sequentially in coordinate order. To determine edges, we analyze node coordinates and line positions. Using this information, we construct a directed acyclic graph (DAG).
 
Using this graph representation of LD, we represent LD in both JSON and metaprogram formats. In the JSON format, we represent the LD as a dictionary; each node ID in the graph is a key, and its attributes and outgoing node IDs form the values. For the metaprogram (code) format, we represent the LD in Python syntax using NetworkX. We traverse the graph starting from the smallest node ID, visiting each node and its successors. Upon first visiting a node, we append a \texttt{G.add\_node(...)} statement with its attributes to the code. Similarly, we append a \texttt{G.add\_edge(...)} statement to the code for each neighbor relationship. This process continues until all nodes from the graph are visited. The generated code can be executed to reconstruct the original graph.

\subsection{Evaluation Metrics}\label{sec:eval_metric}
While text-based programming languages such as Python are typically assessed using unit tests~\cite{chen2021evaluating,austin2021programsynthesislargelanguage, chen2023codet}, VPLs often depend on external simulators~\cite{ray2017survey, ren2024infiniteworldunifiedscalablesimulation}, making consistent automated evaluation challenging across varying environments. To address this, we propose a graph-based automatic evaluation method. We first transform LD programs into NetworkX graphs as described in \S\ref{sec:conversion}. For each LD program, we obtain a ground-truth graph $G = (V, E)$ and a predicted graph $\hat{G} = (\hat{V}, \hat{E})$, which are then compared from two perspectives: partial correctness and exact match.

\textbf{Partial correctness} is measured using F1 scores:
\[
\text{Node F1} = \frac{2\,|V \cap \hat{V}|}{|V| + |\hat{V}|}, \quad
\text{Edge F1} = \frac{2\,|E \cap \hat{E}|}{|E| + |\hat{E}|}
\]
Here, true positives, false positives, and false negatives are computed based on set overlaps. A node (or edge) is considered a match only if all required attributes (e.g., names, types) are exactly equal.

\textbf{Exact Match (EM)} scores assess complete correctness: Node EM is defined as $\mathbf{1}\{\hat{V} = V\}$, Edge EM as $\mathbf{1}\{\hat{E} = E\}$, and Program EM as $\mathbf{1}\{\hat{V} = V \text{ and } \hat{E} = E\}$. Program EM equals 1 only when both the node and edge sets exactly match those of the ground-truth, meaning that the predicted program is identical to it. For a concrete example of metric computation, refer to Appendix~\ref{sec:metric_example}.

\subsection{Results}\label{sec:sft_res}
\paragraph{Training-based method outperforms prompting-based method}\label{par:sft_rag}
To evaluate the effectiveness of the training-based approach compared to the prompting-based approach for industrial VPL generation, we conduct a comparison between the supervised fine-tuning (SFT) method and Retrieval-Augmented Generation (RAG). Among prompting-based methods, we select RAG because this approach has been widely used in prior studies on VPL generation~\cite{xue2024comfybenchbenchmarkingllmbasedagents,zhang2024benchmarking}, and retrieval can provide examples that reflect the domain-specific configurations relevant to generation. SFT uses Llama3.1-8B-Instruct model as the backbone, while RAG uses Llama-3.1-70B-Instruct\footnote{Due to computational constraints, we employed the AWQ-quantized~\cite{MLSYS2024_42a452cb} model: \href{https://huggingface.co/hugging-quants/Meta-Llama-3.1-70B-Instruct-AWQ-INT4}{hugging-quants/Meta-Llama-3.1-70B-Instruct-AWQ-INT4}}. The RAG approach utilizes BM25~\cite{robertson2009probabilistic} to retrieve \(N\) similar prompt-code pairs from the SFT training dataset. We then append them to the input to generate code for the test prompt\footnote{We conducted our experiments using the dataset described in \S\ref{sec:exp_setup}.}. To assess the model’s internal capabilities, we also consider the $N=0$ setting (i.e., zero-shot).  However, since the prompt template does not contain a concrete example of an LD in text format, most of the generated LDs are ill-formed. Therefore, we provide a fixed prompt-code pair even for $N=0$ to inform the model of the expected format.

As shown in Figure~\ref{fig:rag_sft_performance}, SFT consistently outperforms RAG across different numbers of retrieved examples (1, 3, 5, 7, and 9), despite its smaller backbone. Although RAG's performance improves as \(N\) increases, it peaks at $N=5$ and degrades thereafter, suggesting that learning domain-specific configurations in industrial VPL from retrieved examples alone is challenging. For the zero-shot setting, the model performs poorly, with a Node F1 score of just 1.2. This is likely because domain-specific LD configurations, which vary significantly between environments, are not present in the LLM’s pretraining data.

Overall, these findings demonstrate that SFT is a more effective approach in this setting. We observe similar trends with a different LLM (Appendix~\ref{sec:detailed_results}), and the prompt template for the RAG model is provided in Appendix~\ref{sec:rag_details}. 

\paragraph{Training-based method excels across text formats}\label{par:sft_text_format}
To investigate whether the prior results are consistent across different text formats, we extend our experiments to include JSON and the metaprogram formats. We compare SFT with RAG, where performance is reported based on the number of retrieved examples that achieves the highest Program EM for each text format, as the optimal number of examples can vary across formats. 

From the results in Table~\ref{tab:text_format}, we can observe the following: (1) While SFT shows stable performance across different formats, RAG exhibits performance differences in Node/Edge F1, with the metaprogram-based format showing better performance. This aligns with prior results on prompting-based approaches from \citet{xue2024comfybenchbenchmarkingllmbasedagents}. (2) SFT outperforms RAG across all formats. These results further validate the effectiveness of the training-based approach.

\begin{figure*}[!htb]
    \centering
    \includegraphics[width=\textwidth,keepaspectratio]{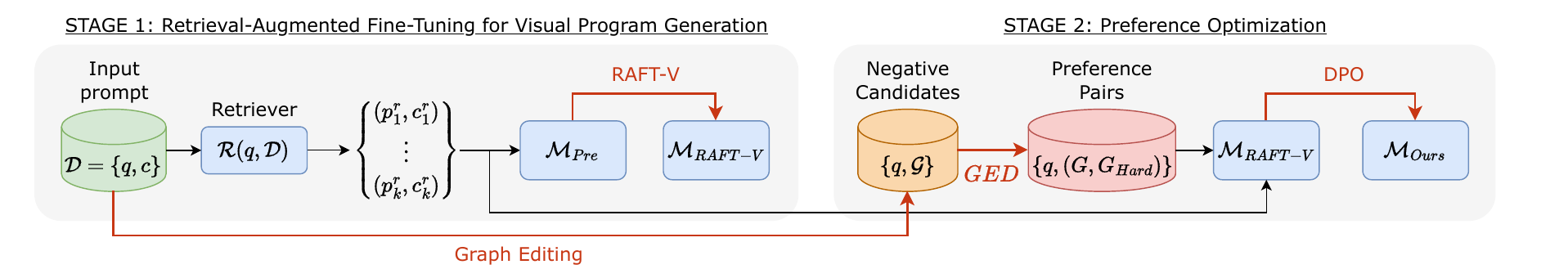}
    \caption{An overview of the two-stage training method. (1) RAFT-V: An off-the-shelf retriever is utilized for relevant prompt augmentation, and training is conducted with cross-entropy loss. (2) Preference Optimization: Preference learning leverages graph-edited preference pairs, with retrieved prompt-code pairs as additional input.}
    \label{fig:main_figure}
\end{figure*}

\section{Proposed Methodology}
In this section, we present a two-stage training strategy to further improve the accuracy of VPL generation. Figure~\ref{fig:main_figure} illustrates an overview of our proposed methodology.

\subsection{Stage 1: Retrieval-augmented Fine-Tuning for Visual Program Generation (RAFT-V)}
Visual programming languages often contain recurring subroutines and modules that are reused in similar contexts~\cite{Terra_Neves_2021}. To leverage this property, we retrieve relevant examples from a training data pool and use them to guide the generation process. Inspired by \citet{zhang2024raft}, which incorporates related documents as context during fine-tuning to improve domain-specific question answering, we extend this idea to the generation of VPL code. Specifically, we retrieve relevant prompt-code pairs as context to enhance the model’s performance.
Let $\mathcal{D}$ be a training dataset consisting of \(N\) prompt-code pairs, where $p$ is a \textit{prompt} and $c$ 
is a corresponding VPL \textit{code}. Given a new input prompt \(q\), we aim to generate the target code \(c_q\). We augment the input to the model with additional prompt-code pairs that are most similar to \(q\). This process is driven by a \textit{retriever} \(\mathcal{R}\), which identifies similar prompts from \(\mathcal{D}\).  \(\mathcal{R}\) ranks all prompts in \(\mathcal{D}\) based on their similarity to \(q\). We then select the top-\(k\) most similar prompts:
\begin{equation}
    \{(p_{i}^r, c_{i}^r)\}_{i=1}^{k} = \mathcal{R}(q, \mathcal{D})
\end{equation}
where \(p_i^r\) and \(c_i^r\) denote the retrieved prompt and code, respectively. We then concatenate the original prompt \(q\) with each retrieved pair \((p_r^i, c_r^i)\) to form the augmented input $q_a$ for the model:
\begin{equation}
    q_a =\bigl(q, [(p_1^r, c_1^r), (p_2^r, c_2^r), \dots, (p_k^r, c_k^r)]\bigr)
\end{equation}
During fine-tuning, the model takes the $q_a$ and learns to generate the target code \(c_q\) from $q_a$ via SFT. At inference, the trained model predicts the code \(\hat{c_q}\) from $q_a$.

\subsection{Stage 2: Preference Optimization}
Preference optimization techniques, such as DPO~\citep{rafailov2023direct}, are widely used as a post-training step following supervised learning in reasoning tasks like code generation~\cite{hui2024qwen25codertechnicalreport, 10.5555/3600270.3601819, weyssow2024codeultrafeedbackllmasajudgedatasetaligning}. These methods rely on high-quality labels to construct preference pairs~\citep{pace2024westofnsyntheticpreferencesselfimproving}.
However, the available preference pairs are virtually nonexistent in domain-specific applications, particularly in underexplored areas such as VPLs. To mitigate this limitation, we propose a graph editing approach to construct preference pairs.

\begin{figure}[t]
\centering
\begin{minipage}{\columnwidth}
\begin{algorithm}[H]
\small
\caption{Graph Editing}\label{alg:graph_augmentation}
\begin{algorithmic}[1]
\STATE \textbf{Input:} Random seed set $S$ with $|S|=s$,  Deletion ratio $\tau$, Graph $G = (V, E)$ with $|V(G)| = n$ nodes and $|E(G)| = m$ edges
\STATE \textcolor[rgb]{0.5,0.5,0.5}{// Initialize the graph}
\STATE $G' \gets G$, $\mathcal{G} \gets \emptyset$;

\STATE \textcolor[rgb]{0.5,0.5,0.5}{// Compute number of nodes to delete}
\STATE $k \gets \lfloor \tau \cdot |V(G')| \rfloor$;

\FOR{each \( i \) in \( S \)}
    \IF{$k > 0$}
        \STATE \textcolor[rgb]{0.5,0.5,0.5}{// Randomly select and reconnect nodes}
        \STATE $\mathcal{X}$ $\gets$ random $k$ nodes from $G'$;
        \STATE Let $\mathcal{W} \to \mathcal{X}  \to \mathcal{Y}$ be a path in $G'$;
        \STATE Connect $\mathcal{W} \to \mathcal{Y}$, preserving the connectivity;

        \STATE \textcolor[rgb]{0.5,0.5,0.5}{// Remove selected nodes}
        \STATE Remove node $\mathcal{X}$ from $G'$;
    \ELSE
        \STATE \textcolor[rgb]{0.5,0.5,0.5}{// If no node is removed, copy a random node}\label{alg:add}
        \STATE Copy a random node-edge pair $(v_{\text{copy}}, e_{\text{copy}}) \in G'$;
        \STATE Connect a randomly selected node $v_{\text{orig}}$ in $G'$ to $v_{\text{copy}}$ using $e_{\text{copy}}$;
    \ENDIF
    \STATE \textcolor[rgb]{0.5,0.5,0.5}{// Store the modified graph}
    \STATE Add $G'$ to set $\mathcal{G}$;
\ENDFOR
\STATE Return \( \mathcal{G} = \{ G'_1, G'_2, \dots, G'_s \} \)
\end{algorithmic}
\end{algorithm}

\end{minipage}
\end{figure}
\paragraph{Negative sampling via graph editing} 
As described in \S\ref{sec:conversion}, we transform visually structured programs into NetworkX graphs to facilitate access to nodes and edges. Based on this graphical representation, we introduce a method for collecting preference pairs. We collect negative samples through graph editing, as described in Algorithm~\ref{alg:graph_augmentation}.
Given a graph $G$ corresponding to ground-truth VPL code, we randomly delete nodes and edges, with a deletion ratio $\tau$ set to 0.1. This modifies approximately 10\% of the graph structure. For graphs with fewer than 10 nodes, no deletions are performed. Instead, as shown in line~\ref{alg:add}, we augment the graph by duplicating and adding a node-edge pair. To ensure diversity, we generate 10 negative samples for each graph using different random seeds. The effect of varying $\tau$ is analyzed in \S\ref{sec:tau}.

\paragraph{Hard negative selection}
Preference optimization using hard negative samples is beneficial for learning subtle differences that supervised learning alone may not capture~\cite{zhu2024selfsupervised, chen2024on}.
This suggests that the difference in connectivity between hub nodes and leaf nodes is substantial, and when nodes are randomly deleted, the quality of negative data varies considerably.
Therefore, we select hard negative samples from the negative set collected via graph editing. To do so, we utilize the Graph Edit Distance (GED)~\cite{6313167, abu2015exact}, which quantifies the structural difference between graphs. Specifically, we use the GED-based metric for selection:
\begin{equation}
    \text{GED}(G,G')=\min_{T\in\mathcal{T}(G,G')}\sum\limits_{e \in T}\text{cost}(e)
\end{equation}
where $\mathcal{T}(G,G')$ is the set of all possible edit operation that transform $G'$ into $G$. Given $\mathcal{T}(G,G')$ and the cost, the GED is the minimum associative cost of each edit operation across all possible edit paths. In this study, the cost of an edit path is measured using the Levenshtein distance~\citep{black1998dictionary}. To construct preference pairs, we select the graph with the lowest edit distance to the reference graph from $\mathcal{G}=\{G'_1\dots G'_s\}$ collected through multiple seeds. We designate the graph with the lowest score as the hard negative $G_{\text{Hard}}$. Finally, we construct a preference pair by pairing ground-truth graph $G$ with $G_{\text{Hard}}$. 

\paragraph{Direct preference optimization}
Preference learning is conducted using the selected preference pairs, with the retrieved prompt and its corresponding VPL code augmented as input. Implementation details are provided in Appendix~\ref{sec:implementation_details}.
\begin{table*}[ht]
\tiny
\centering
\resizebox{.9\textwidth}{!}{%
\begin{tabular}{llccccc}
\toprule
Format & Method & \multicolumn{1}{l}{Node F1} & \multicolumn{1}{l}{Edge F1} & \multicolumn{1}{l}{Node EM} & \multicolumn{1}{l}{Edge EM} & \multicolumn{1}{l}{Program EM} \\ \midrule
 & SFT & 84.8 / 79.0 & 74.7 / 68.0 & 55.2 / 46.4 & 54.8 / 45.6 & 54.8 / 45.6 \\
 & RAFT-V & 88.1 / \textbf{87.4} & 80.9 / 78.7 & 63.8 / 59.6 & 62.8 / 58.8 & 62.8 / 58.8 \\
\multirow{-3}{*}{XML} & \cellcolor[HTML]{EFEFEF}Ours & \cellcolor[HTML]{EFEFEF}\textbf{89.6 / 87.4} & \cellcolor[HTML]{EFEFEF}\textbf{82.6 / 79.0} & \cellcolor[HTML]{EFEFEF}\textbf{66.2 / 61.2} & \cellcolor[HTML]{EFEFEF}\textbf{65.2 / 60.0} & \cellcolor[HTML]{EFEFEF}\textbf{65.2 / 60.0} \\ \midrule
 & SFT & 85.1 / 81.2 & 74.9 / 69.3 & 53.6 / 46.0 & 52.6 / 45.4 & 52.6 / 45.4 \\
 & RAFT-V & 90.1 / 88.4 & 82.0 / 79.6 & 63.8 / 59.2 & 63.2 / 57.4 & 63.2 / 57.4 \\
\multirow{-3}{*}{JSON} & \cellcolor[HTML]{EFEFEF}Ours & \cellcolor[HTML]{EFEFEF}\textbf{90.7 / 88.7} & \cellcolor[HTML]{EFEFEF}\textbf{83.0 / 80.5} & \cellcolor[HTML]{EFEFEF}\textbf{66.0 / 61.6} & \cellcolor[HTML]{EFEFEF}\textbf{65.2 / 60.2} & \cellcolor[HTML]{EFEFEF}\textbf{65.2 / 60.2} \\ \midrule
 & SFT & 86.3 / 81.7 & 75.7 / 70.2 & 55.2 / 47.6 & 54.0 / 47.2 & 54.0 / 47.2 \\
 & RAFT-V & 89.9 / 89.1 & 82.4 / 80.4 & 64.6 / 60.6 & 63.8 / 59.6 & 63.8 / 59.6 \\
\multirow{-3}{*}{Metaprogram} & \cellcolor[HTML]{EFEFEF}Ours & \cellcolor[HTML]{EFEFEF}\textbf{90.6 / 89.7} & \cellcolor[HTML]{EFEFEF}\textbf{83.6 / 81.6} & \cellcolor[HTML]{EFEFEF}\textbf{68.6 / 63.2} & \cellcolor[HTML]{EFEFEF}\textbf{67.2 / 61.8} & \cellcolor[HTML]{EFEFEF}\textbf{67.2 / 61.8} \\ \bottomrule
\end{tabular}%
}
\caption{Main results of the 2-stage training strategy. Each score is presented in the order of Llama-3.1-8B-Instruct / Qwen2.5-7B-Instruct.}
\label{tab:main_result}
\end{table*}

\section{Experimental Settings}\label{sec:exp_setup}
\paragraph{Dataset} Due to the lack of publicly available datasets, we created our own by annotating ladder diagrams from actual production environments. Using XG5000~\cite{XG5000Manual}, we exported these diagrams as XML files and divided them into functional units, where each unit consists of one or more interconnected rungs designed to perform a specific function. An experienced PLC programmer then annotated natural language instructions for each unit. The dataset was randomly split into training, validation, and test sets of 13,124, 500, and 500 samples. 80\% of the training data was used for SFT, while the remaining 20\% was used for preference learning.
Appendix~\ref{sec:data_samples} shows detailed information about the dataset utilized in our study.

\paragraph{Implementation details} We utilize Llama-3.1-8B-Instruct~\cite{dubey2024llama} as the main backbone model and also use Qwen2.5-7B-Instruct~\cite{yang2024qwen2} to assess the generalizability of our method. To facilitate task understanding, we provide a detailed task description and explanations of the visual elements used in ladder diagrams via the system prompt. For models using retrieval, we utilize BM25~\cite{robertson2009probabilistic}, which is a widely used lexical matching-based method. Using BM25, we augment the input to these retrieval-based models with a top-1 retrieved prompt-code pair from the training dataset\footnote{System prompts are in Appendix~\ref{sec:system_prompt}, and further implementation details are in Appendix~\ref{sec:implementation_details}.}.
\section{\label{sec:overall_performance}Main Results}
\paragraph{Stage 1 (RAFT-V) improves upon SFT} Table~\ref{tab:main_result} compares the performance of SFT, RAFT-V, and our two-stage method across different output formats (XML, JSON, metaprogram) with two backbone models. By incorporating retrieval augmentation, RAFT-V consistently outperforms SFT and significantly enhances VPL generation. For example, in the JSON format, RAFT-V raises the Program EM from 52.6\% to 63.2\%, demonstrating more accurate VPL generation. Similar improvements are observed for other metrics.

\paragraph{Preference optimization (Stage 2) yields further gains} Building on Stage 1's improvements, Stage 2 further boosts correctness through preference optimization, particularly in terms of exact match (EM) scores. As shown in Table~\ref{tab:main_result}, our two-stage approach improves Program EM over RAFT-V by 3.4\% in the metaprogram format and achieves an overall 13.2\% gain compared to SFT. Because EM only assigns a score when the entire graph exactly matches, even minor generation failures can significantly impact the metric. These results show that preference optimization reduces such minor generation errors and ensures more precise outputs. Notably, these gains are consistent across all output formats and base models, demonstrating the robustness of our approach.\footnote{See Appendix~\ref{sec:human_eval} for human evaluation details.} Since the generated PLC code is directly deployed in factory environments, careful review by programmers is still essential. In this context, a more than 10\% increase in Program EM scores across all output formats compared to SFT is practically significant in real industrial settings. This translates to faster validation cycles and reduced manual effort, making the overall development process more efficient and cost-effective.

\section{\label{sec:result}Discussion}
In this section, we evaluate RAFT-V under different retrieved examples (\S\ref{sec:retrieved_examples}), analyze the deletion ratio in graph editing (\S\ref{sec:tau}), compare graph sampling based on graph editing to an existing sampling strategy (\S\ref{sec:data_augmentation}), and investigate the impact of program complexity on our approach (\S\ref{sec:difficulty}).

\subsection{\label{sec:retrieved_examples}Impact of Number of Retrieved Examples}
\begin{table}[!t]
\centering
\resizebox{\columnwidth}{!}{%
\begin{tabular}{l|ccccc}
\toprule
\( k \) & Node F1 & Edge F1 & Node EM & Edge EM & Program EM \\ \midrule
\rowcolor[HTML]{EFEFEF}
1 (Ours)      & 89.9    & 82.4    & 64.6    & 63.8    & 63.8     \\
2       & 90.6    & 83.5    & \textbf{66.4} & \textbf{65.6} & \textbf{65.6} \\
3       & \textbf{90.9} & \textbf{84.0} & 65.4    & 64.6    & 64.6     \\ \bottomrule
\end{tabular}%
}
\caption{Performance of RAFT-V across different numbers of retrieved examples $k$. We use the Llama3.1-8B-Instruct model with metaprogram format.}
\label{tab:retrieval_quantity_performance}
\end{table}
To assess whether the number of retrieved examples affects performance, we evaluate RAFT-V trained with different numbers of retrieval examples \( k \in \{1, 2, 3\} \) (Table~\ref{tab:retrieval_quantity_performance}). Although increasing retrieval examples \(k\) generally improves generation quality, the gains are marginal, which indicates that even a few examples are able to capture functional patterns in visual programming languages. Based on this observation, we set \( k = 1 \) for our main experiments.

\subsection{\label{sec:tau}Variation in Deletion Ratio}
\begin{table}[!h]
\centering
\resizebox{\columnwidth}{!}{%
\begin{tabular}{l|ccccc}
\toprule
$\boldsymbol{\tau}$ & Node F1 & Edge F1 & Node EM & Edge EM & Program EM \\ \midrule
0 & \textbf{90.7} & 83.4 & 67.2 & 65.8 & 65.8 \\
\rowcolor[HTML]{EFEFEF}
0.1 (Ours) & 90.6 & \textbf{83.6} & \textbf{68.6} & \textbf{67.2} & \textbf{67.2} \\
0.5 & 90.5 & 83.3 & 68.2 & 66.8 & 66.8 \\
0.9 & 90.6 & 83.4 & 67.4 & 66.0 & 66.0 \\ \bottomrule
\end{tabular}%
}
\caption{Impact of varying deletion ratios. Results for Llama-3.1-8B-Instruct model using metaprogram format.}
\label{tab:dpo_tau}
\end{table}
Furthermore, we report the performance variation depending on the deletion ratio $\tau$. We train the model using preference pairs with different $\tau$ values, where $\tau \in \{0, 0.1, 0.5, 0.9\}$. Table~\ref{tab:dpo_tau} shows that F1 scores remain stable across $\tau$ values, while EM scores decrease as $\tau$ increases. In particular, when $\tau$ is 0.9, the negative samples during preference training are graphs with 90\% of the original graph removed. As a result, the model can easily distinguish them, making the training less effective. Based on these results, we set $\tau$ to 0.1.

\begin{table}[!ht]
\centering
\resizebox{\columnwidth}{!}{%
\begin{tabular}{l|ccccc}
\toprule
Methods & Node F1 & Edge F1 & Node EM & Edge EM & Program EM \\ \midrule
RAFT-V (100\%) & 90.6 & \textbf{83.6} & 66.6 & 65.2 & 65.2 \\
\rowcolor[HTML]{EFEFEF}
RAFT-V & 89.9 & 82.4 & 64.6 & 63.8 & 63.8 \\ \midrule
BoN (Random) & 90.4 & 83.2 & 66.8 & 65.6 & 65.6 \\
BoN (Unstrict) & \textbf{90.7} & 83.5 & 67.0 & 65.8 & 65.8 \\
BoN (Strict) & 90.5 & 83.3 & 67.4 & 66.2 & 66.2 \\
Ours (Random) & 90.6 & 83.5 & 68.0 & 66.6 & 66.6 \\
\rowcolor[HTML]{EFEFEF}
Ours & 90.6 & \textbf{83.6} & \textbf{68.6} & \textbf{67.2} & \textbf{67.2} \\ \bottomrule
\end{tabular}%
}
\caption{Comparison of BoN sampling and graph editing. Results for metaprogram using Llama-3.1-8B-Instruct.}
\label{tab:dpo_aug}
\end{table}
\subsection{\label{sec:data_augmentation}Analysis of Graph Editing}
We validate our graph editing approach for collecting preference pairs by comparing it with Best-of-N (BoN) sampling\footnote{For BoN sampling, we use nucleus decoding~\citep{Holtzman2020The} with temperature=1.0, top\_p=1.0.}~\citep{10.5555/3495724.3495977, snell2025scaling}, which selects the best generation among $N$ candidates.
We generate $N=10$ outputs from the RAFT-V model and compare them with our negative candidate set $\mathcal{G}=\{G'_1\dots G'_{10}\}$, which is derived from graph editing (algorithm~\ref{alg:graph_augmentation}).

\paragraph{Preference learning is effective}
To address the concern that the gains in the 2-stage approach may result from the additional data used for preference learning, we compare it with a 1-stage approach, where the model is trained on all the data without a second stage and is referred to as \underline{RAFT-V (100\%)}. As shown in Table~\ref{tab:dpo_aug}, RAFT-V improves performance as the dataset size increases. However, RAFT-V (100\%) shows only marginal improvement in EM scores compared to preference-learned models. Although RAFT-V (100\%) demonstrates effectiveness in improving F1 scores when compared to BoN-based preference learning baselines, it is insufficient to minimize minor errors (EM).

\paragraph{Editing-based negative selection is efficient}
We introduce four baselines for comparison with graph editing: \underline{BoN (Random)}, where $\mathcal{G}$ is sampled from BoN, and $G_{\text{Hard}}$ is randomly selected from BoN-sampled $\mathcal{G}$; \underline{BoN (Unstrict)}, which selects preference pairs based on GED; and \underline{BoN (Strict)}, which considers preference pairs for training only when the sampled output includes an exact match with the correct answer (i.e., GED = 0).  Finally, in the case of \underline{Ours (Random)}, $\mathcal{G}$ is sampled from graph editing, but $G_{\text{Hard}}$ is randomly selected instead of using GED.

As shown in Table~\ref{tab:dpo_aug}, BoN (Strict) constructs preference pairs based on exact matches and achieves the highest accuracy among BoN-based methods despite utilizing only 30\% of the data. This result demonstrates that dataset quality has a more significant impact on preference learning than dataset size~\cite{hou2024doesrlhfscaleexploring, kim2024aligninglargelanguagemodels}. However, BoN sampling often fails to generate challenging cases consistently, as negative samples are selected from the sampled outputs. In contrast, our editing-based negative selection provides a more systematic approach to generating negative samples, which consistently enables the generation of hard negative pairs with higher efficiency. 

\begin{figure}[!t]
    \centering
    \includegraphics[width=\columnwidth,keepaspectratio]{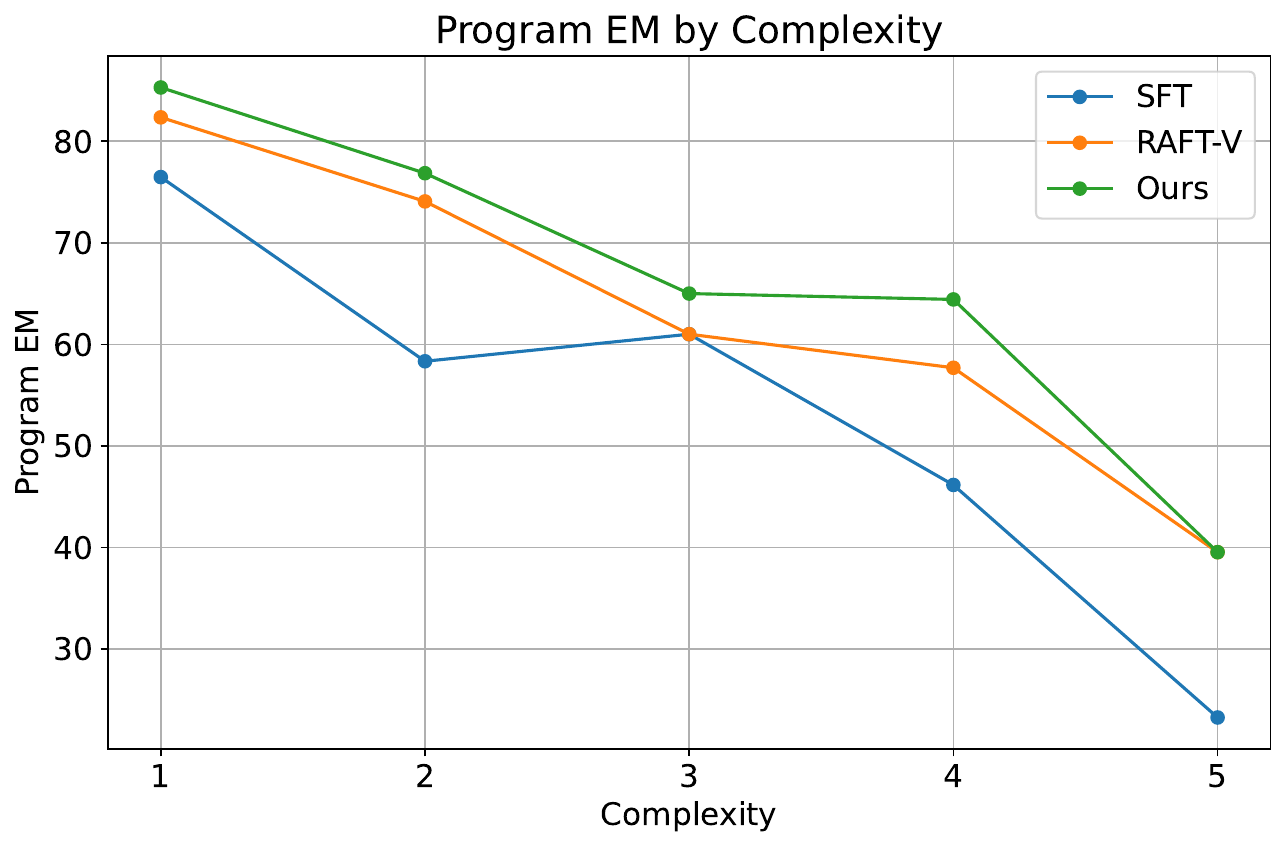}
    \caption{Program EM score across different complexities. We use the metaprogram format with the Llama3.1-8B-Instruct model.}
    \label{fig:total_EM_trend_complexity}
\end{figure}
\subsection{\label{sec:difficulty}Performance Across Program Complexity}
To evaluate our methodology's performance across varying program complexities, we convert each program (VPL code) in the test split into NetworkX graphs. We define complexity as the total number of nodes and edges in the graph. We then sort programs by complexity and split them into five percentile ranges: 0–20\%, 20–40\%, 40–60\%, 60–80\%, and 80–100\%, labeling them from 1 to 5. Figure~\ref{fig:total_EM_trend_complexity} shows the average Program EM across these categories. Our approach consistently outperforms the SFT baseline, widening the gap at higher complexities (+18.3\% in 4, +16.3\% in 5). These findings highlight our method's robustness, with benefits increasing in challenging scenarios.
\section{Related Work}
\paragraph{LLMs for Visual Program Generation}
Visual programming systems (e.g., LabView~\cite{bitter2006labview}, XG5000~\cite{XG5000Manual}) typically feature node-based interfaces that let users visually write and modify programs. Recently, researchers have begun utilizing LLMs to generate VPLs, as they are known for their powerful text-based code generation capabilities. For example, \citet{cai-etal-2024-low-code} integrates low-code visual programming with LLM-based task execution for direct interaction with LLMs, while \citet{zhang2024benchmarking} studies generation of node-based visual dataflow languages in audio programming. Similarly, \citet{xue2024comfybenchbenchmarkingllmbasedagents,52868} investigates Machine Learning workflow generation from natural language commands and demonstrates that metaprogram-based text formats outperform other formats like JSON. However, these prompting-based methods face limitations for VPLs like Ladder Diagram, where custom I/O mapping and domain-specific syntax are crucial. Thus, we study fine-tuning approaches with domain-specific data to better capture these details.

\paragraph{LLM-based PLC code generation}
Programmable Logic Controllers (PLCs) are essential components in industrial automation and are used to control machinery and processes reliably and efficiently. Among the programming languages defined by the IEC 61131-3 standard~\cite{IEC61131-3}, Structured Text (ST) and Ladder Diagram (LD) are commonly used for programming PLCs. Research in this area has focused on utilizing LLMs to generate ST code from natural language descriptions. Recent studies have demonstrated the potential of LLMs in generating high-quality ST code~\cite{koziolek2023chatgpt, koziolek2024llm}, enhancing safety and accuracy with verification tools and user feedback~\cite{fakih2024llm4plc}, and automating code generation and verification using multi-agent frameworks~\cite{liu2024agents4plc}. Although these advances have improved PLC code generation, they primarily focus on ST, despite LD being widely used in industrial settings due to its similarity to electrical circuits~\cite{ladderlogic}. While \citet{Zhang_2024} attempts to generate LD as an ASCII art based on user instructions in a zero-shot manner, their findings show that even advanced LLMs struggle with basic LD generation. These limitations highlight the necessity of training-based methods for LD generation. In this work, we address this gap by introducing a training-based approach for LLMs to generate LD and thus pave the way for the broader adoption of AI-assisted PLC programming.
\section{Conclusion}
In this paper, we have demonstrated the importance of fine-tuning for VPL generation and introduced a novel two-stage training strategy. By combining retrieval augmentation leveraging repetitive subroutines in VPLs with preference learning through graph editing, we achieved significant improvements in LD generation. Our work marks a crucial step toward LLM-based LD generation, and given that the method leverages VPL characteristics, it holds potential for broader applicability. 

\section*{Limitations}
One limitation of our study is that the experiments were conducted exclusively on ladder diagrams. Although ladder diagrams are widely used in visual programming languages (VPLs), extending our methodology to other VPLs is necessary for broader applicability. Nonetheless, ladder diagrams remain crucial in industrial automation but have been largely overlooked due to the text-centric design of most large language models (LLMs). Prior studies have struggled to generate even basic ladder diagrams, highlighting a significant gap. To bridge this, we proposed a method to represent ladder diagrams in multiple text formats, enabling LLMs to process them as a graphical language. This approach addresses the limitations of previous studies and opens new possibilities for LLM-based generation of visual programming languages.

Another limitation is that our dataset cannot be publicly released due to strict confidentiality constraints, as it contains proprietary industrial processes used in manufacturing. To safeguard sensitive information, we have anonymized and provided only a partial subset. Our dataset adhered to the IEC 61131-3 international standard and specifically utilized ladder diagrams, which are a widely adopted language in PLC programming. While our dataset has unique attributes for ladder elements and rung structures that differ across users, it remains applicable to other ladder diagram-based programs that adhere to the IEC 61131-3 standard. Consequently, our methodology is not restricted to this particular dataset but rather is applicable to other industrial formats, which demonstrates scalability and adaptability. Yet, recognizing the importance of further generalization, we acknowledge the need for a publicly available VPL-based benchmark, which we leave as future work.
\section*{Ethical Considerations}
In our research, we employed models such as Llama-3.1-8B-Instruct and Meta-Llama-3.1-70B-Instruct-AWQ-INT4, both of which are released under the Llama 3.1 Community License, as well as Qwen2.5-7B-Instruct, released under the Apache License 2.0, and Qwen2.5-72B-Instruct-AWQ, released under the Qwen license. All models were used strictly for research purposes, and no artifacts were utilized beyond the scope of the study.

\section*{Acknowledgments}
This work was supported by Hyundai Mobis (47.5\%). This research was supported by the MSIT(Ministry of Science and ICT), Korea, under the ITRC(Information Technology Research Center) support program(IITP-2025-2020-0-01789) supervised by the IITP(Institute for Information \& Communications Technology Planning \& Evaluation, 47.5\%). This work was supported by Institute of Information \& communications Technology Planning \& Evaluation (IITP) grant funded by the Korea government(MSIT) (No.RS-2019-II191906, Artificial Intelligence Graduate School Program(POSTECH), 5\%).


\bibliography{custom}

\onecolumn
\appendix

\section{\label{sec:metric_example}An Example of Metric Evaluations}
\begin{figure*}[!htbp]
    \centering
    \includegraphics[width=0.9\textwidth]{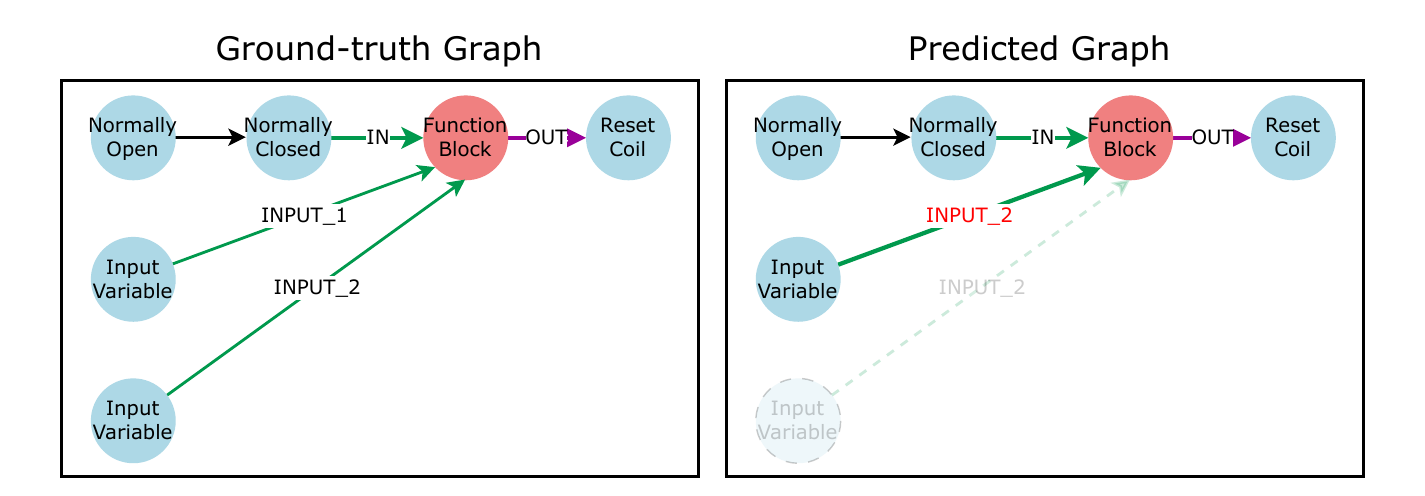}
    \caption{Example illustrating the evaluation metrics. Node and edge attributes have been modified from the original data due to security concerns.}
    \label{fig:metric_v2}
\end{figure*}
Figure~\ref{fig:metric_v2} presents the ground-truth and predicted graphs used for the following evaluation example. Based on these graphs, we illustrate how the proposed metrics capture both partial and exact correctness of the predicted Ladder Diagram (LD) program.
First, consider the \textbf{node-level evaluation}. The ground-truth graph $G = (V, E)$ contains 6 nodes, while the predicted graph $\hat{G} = (\hat{V}, \hat{E})$ contains 5 nodes. All 5 predicted nodes correctly match with nodes in $V$, including all required attributes such as type and name. Hence,
\[
\text{Precision}_N = \frac{|V \cap \hat{V}|}{|\hat{V}|} = \frac{5}{5} = 1.0, \quad
\text{Recall}_N = \frac{|V \cap \hat{V}|}{|V|} = \frac{5}{6} = 0.833
\]
Using these values, the Node F1 score is:
\[
\text{Node F1} = \frac{2 \times 1.0 \times 0.833}{1.0 + 0.833} = 0.91
\]
Next, we examine the \textbf{edge-level evaluation}. The reference graph has 5 edges, while the predicted graph contains only 4. Among the predicted edges, 3 match the ground-truth edges exactly in terms of connected nodes and attributes. Therefore,
\[
\text{Precision}_E = \frac{3}{4} = 0.75, \quad
\text{Recall}_E = \frac{3}{5} = 0.6
\]
\[
\text{Edge F1} = \frac{2 \times 0.75 \times 0.6}{0.75 + 0.6} = 0.67
\]
Finally, we assess \textbf{Exact Match (EM)}. Since both the node and edge sets do not exactly match those in the reference graph—one node is missing and one edge is incorrect—all EM scores are 0:
\[
\text{Node EM} = 0, \quad \text{Edge EM} = 0, \quad \text{Program EM} = 0
\]

\section{Further Implementation Details}\label{sec:implementation_details}
\paragraph{Supervised fine-tuning (including RAFT-V)}
We fine-tuned models using SFT with 10 epochs, a batch size of 8, and a learning rate of $5 \times 10^{-5}$. We applied LoRA~\cite{hulora} with a rank of 256, $\alpha = 256$, and a dropout rate of 0.05. LoRA adaptation was applied to the following target modules: q\_proj, v\_proj, k\_proj, o\_proj. Using the AdamW~\cite{loshchilov2018decoupled} optimizer, we minimized the cross-entropy loss $\mathcal{L}_{\text{CE}}$ between ground-truth code $c$ and the predicted code $\hat{c}$ as follows:
\begin{equation}
\nonumber
    \mathcal{L}_{\text{CE}} = - \sum_{t=1}^{T} \sum_{v \in V} c_t(v) \log \hat{c}_t(v)
\end{equation}
where $T$ is the sequence length and $V$ is a vocabulary size. The SFT process took 6 hours using 4 A100-80GB GPUs, while the SFT process for RAFT-V took 8 hours using the same hardware configuration.

\paragraph{Preference learning} For preference learning, we utilized Direct Preference Optimization (DPO)~\cite{rafailov2023direct} and trained models for 5 epochs with a batch size of 64. The learning rate was set to $1 \times 10^{-7}$, with a warmup ratio of 0.03, a weight decay of 0.01, and $\beta=0.1$. As with SFT, {LoRA} was applied with the same rank, $\alpha$, dropout rate, and target modules.

Given a pair of responses, a preferred response $y^+$ and a dispreferred response $y^-$ for a given input $x$, we minimize the following DPO loss:

\begin{equation}
\nonumber
    \mathcal{L}_{\text{DPO}}(\theta) = -\mathbb{E}_{(x,y^+, y^-) \sim \mathcal{D}} \left[ 
    \log \sigma \left( \beta \left( \log \frac{\pi_\theta(y^+ \mid x,\mathcal{R}(x))}{\pi_{\text{ref}}(y^+ \mid x,\mathcal{R}(x))} - \log \frac{\pi_\theta(y^- \mid x,\mathcal{R}(x))}{\pi_{\text{ref}}(y^- \mid x,\mathcal{R}(x))} \right) \right) 
    \right]
\end{equation}
where $\sigma(z)$ is the sigmoid function, $\beta$ is a scaling factor controlling the strength of preference optimization, $\mathcal{R}(x)=(p^r, c^r)$ represents a retrieved prompt and its corresponding VPL code obtained from a retriever $\mathcal{R}$, and $\mathcal{D}$ represents the dataset of preference-labeled samples. The reference model $\pi_{\text{ref}}$ serves as a baseline to prevent reward overoptimization, ensuring stable preference learning. The preference training stage took 6 hours on 4 A100-80GB GPUs.
We measured the loss on the validation set at each epoch in both training stages and applied early stopping based on this criterion, with a patience value of 2. 
\paragraph{Sampling parameters} We employed beam search decoding with a beam size of 4 during inference. 

\section{Dataset Details}\label{sec:data_samples}

\begin{table*}[!htbp]
    \small
    \centering
    \begin{tabular}{|p{0.15\textwidth}|p{0.8\textwidth}|} 
        \toprule
        \textbf{Type} & \textbf{Content} \\ \midrule

        Prompt 
        & \textbf{Program Description:} 셀 이재기 서보 조그 고속 하한 값 인터락 프로그램을 만들어줘. \\[3pt]
        & \textcolor{gray}{(Create a cell transfer servo jog high-speed interlock program.)} \\[3pt]
        & \textbf{Detailed Description:} 조그 고속 속도 설정값이 20 이하일 때 조그 고속속도를 20으로 설정해줘. \\[3pt]
        & \textcolor{gray}{(When the jog high-speed setting value is 20 or less, set the jog high-speed to 20.)} \\[3pt] \midrule

        \makecell[l]{XML \\ (Anonymized)}
        & \begin{lstlisting}[basicstyle=\ttfamily, aboveskip=0pt, belowskip=0pt]
<Rung>
    <Element Attributes..., Coordinate="X"></Element>
    <Element Attributes..., Coordinate="Y"></Element>
    <Element Attributes..., Coordinate="Z"></Element>
</Rung>
\end{lstlisting} \\ \bottomrule
    \end{tabular}
    \caption{Program description and anonymized XML example.}
    \label{tab:appendix_sample}
\end{table*}

\begin{table*}[!htbp]
\centering
\begin{tabular}{l|cc}
\toprule
Process Type & Count & Ratio (\%) \\ \midrule
Cell Distribution & 2,893 & 20.5 \\
Cell Processing & 3,093 & 21.9 \\
Cell Vision Inspection & 726 & 5.1 \\
Cell Stacking & 3,706 & 26.2 \\
Pack End-plate Supply & 2,808 & 19.9 \\
Other Processes & 898 & 6.4 \\
\bottomrule
\end{tabular}
\caption{Distribution of process types in the dataset.}
\label{tab:dataset_statistics}
\end{table*}

An example from the dataset used in this study is provided. The dataset was extracted from ladder logic (LD) programs deployed in real-world battery cell manufacturing facilities, and consists of \textbf{prompt} and \textbf{code} pairs. The prompt was collected in Korean and used without translation for model training and evaluation. The prompt is further divided into \textbf{Program Description}, providing a high-level summary of the functionality, and \textbf{Detailed Description}, specifying task parameters and conditions. The prompts were created by an experienced PLC programmer from the Republic of Korea, ensuring the incorporation of domain expertise into the collected data. The programmer was explicitly informed that the dataset would be collected for model training and evaluation and used strictly for research purposes. We obtained consent prior to data collection. The \textbf{code} is represented in XML format, which describes the ladder diagram with elements. These elements are listed sequentially, and each element includes its attributes (e.g., variable names, parameters) and coordinate information. The datasets were constructed by an experienced PLC engineer, and all sensitive information was anonymized to ensure data confidentiality.

In addition, we report the distribution of process types included in the dataset. We constructed fine-grained and diverse operational data across multiple stages of production, including cell distribution, processing, stacking, vision inspection, and pack component supply. The statistics of the dataset for each stage are presented in Table~\ref{tab:dataset_statistics}. We hope that this statistical information will serve as a valuable reference for future research.
\section{\label{sec:human_eval}Human Evaluation}
To further evaluate the effectiveness of our methods, we conduct human evaluations on a randomly selected 50 test set examples. Due to the dataset's security sensitivity, human evaluations were conducted exclusively by PLC engineers who had authorized access to confidential information. 
Three experienced LD programmers evaluated the generated code based on functionality and assigned scores on a scale of 1 to 5. The evaluation process is as follows: First, we convert the generated ladder program in each text format into an actual Ladder Diagram that can be opened in the IDE (XG5000). Then, we manually capture both the reference and predicted ladder programs from the IDE. Finally, the evaluators are asked to assess each reference–prediction screenshot pair according to the criteria. The criteria for scoring are reported in Table~\ref{tab:human evaluation criteria}. The ratings were based on their professional experience and the practical applicability of the code in real industrial settings.

The results of the human evaluation are shown in Table~\ref{tab:human_evaluation}. Compared to SFT-only, RAFT-V showed an improvement of 0.26 points, while our method outperformed RAFT-V by 0.02 points. Furthermore, we report Fleiss' kappa coefficient~\citep{fleiss1971measuring} to statistically evaluate the level of agreement among human evaluators. The results indicate that our proposed methodology demonstrates the highest degree of inter-rater consistency.

\begin{table*}[!htbp]
\centering
\begin{tabular}{l|l}
\toprule
Score & Assessment Criteria \\\midrule
5 & Immediately usable in real-world deployment. \\
4 & Mostly appropriate; minor polishing may help. \\
3 & Valid but awkward; manual revision is needed. \\
2 & Mostly correct syntax but clearly wrong logic or intent. \\
1 & Unexecutable or severely broken structure with partial syntax validity. \\
\bottomrule
\end{tabular}
\caption{Scoring guidelines for human evaluation.}
\label{tab:human evaluation criteria}
\end{table*}

\begin{table*}[!htbp]
\centering
\begin{tabular}{l|cc}
\toprule
Method        & Functional Score $\uparrow$ & Kappa Score $\uparrow$ \\ \midrule
SFT-only      & 4.34            &       0.62      \\
RAFT-V        & 4.60            &         0.58    \\
Ours & \textbf{4.62} & \textbf{0.66}   \\ \bottomrule
\end{tabular}
\caption{Average of human evaluation results. We used the metaprogram format for evaluation.}
\label{tab:human_evaluation}
\end{table*}
\section{More Details on RAG}\label{sec:rag_details}
The prompt template for RAG models is as follows. The RAG model takes the same system prompt as the training-based model depending on the text format.

\begin{shk}
message = [
    {"role": "system", "content": {system prompt}},
    {"role": "user", "content": retrieved_prompt}, 
    ...
    {"role": "assistant", "content": retrieved_code},
    {"role": "user", "content": {test_prompt}}
]
\end{shk}
\noindent\begin{minipage}{\textwidth}
\captionsetup{type=figure}
\captionof{figure}{Prompt template for RAG models.}
\end{minipage}

Since some outputs of RAG models were ill-formed (particularly in the case of Qwen), we applied postprocessing to refine them. Extra elements such as \verb|```xml|, \verb|```json|, \verb|```python|, or unrelated code snippets were removed. For inference, we employ beam search with beam size of 4.

\section{Detailed Results}\label{sec:detailed_results}
In this section, we present detailed experimental results. Table~\ref{tab:rag_versus_sft_llama} shows the performance comparison between the SFT model and RAG, with a larger LLM within the same family.

\begin{table*}[!htbp]
\tiny
\centering
\resizebox{0.9\textwidth}{!}{%
\begin{tabular}{lcccccc}
\toprule
Format& Method & Node F1& Edge F1& Node EM& Edge EM& Program EM\\ \midrule

\multirow{6}{*}{XML}& SFT& \textbf{84.8} / \textbf{79.1}&\textbf{74.7} / \textbf{68.2} & \textbf{55.2} / \textbf{46.8}& \textbf{54.8} / \textbf{46.0}& \textbf{54.8} / \textbf{46.0}\\
& $N=0$& 1.2 / 1.0& 0.5 / 0.0& 0.0 / 0.0& 0.4 / 0.0& 0.0 / 0.0\\
& $N=1$& 69.0 / 19.5& 58.2 / 13.9& 39.0 / 8.8& 38.6 / 8.6& 38.0 / 8.6\\
& $N=3$& 76.3 / 61.4& 67.6 / 53.3& 49.8 / 40.2& 49.0 / 39.4& 49.0 / 39.4\\
& $N=5$& 74.4 / 60.7& 67.1 / 53.8& 49.6 / 41.2& 49.2 / 40.6& 49.2 / 40.6\\
& $N=7$& 59.3 / 53.6& 53.5 / 47.9& 42.8 / 37.8& 42.2 / 37.6& 42.2 / 37.6\\
& $N=9$& 50.5 / 44.9& 45.1 / 40.4& 37.6 / 34.4& 36.8 / 34.0& 36.8 / 34.0\\ \midrule
\multirow{6}{*}{JSON}& SFT& \textbf{85.1} / \textbf{81.2} & \textbf{74.9} / \textbf{69.3} & \textbf{53.6} / \textbf{46.8}& \textbf{52.6} / 46.4& \textbf{52.6} / 46.4\\
& $N=0$& 0.0 / 0.0& 0.0 / 0.0& 0.0 / 0.0& 0.0 / 0.0& 0.0 / 0.0\\
& $N=1$& 1.2 / 5.3& 0.8 / 3.9& 0.6 / 3.6& 0.6 / 4.0& 1.1 / 3.6\\
& $N=3$ & 79.2 / 73.4& 69.8 / 62.7& 48.2 / 43.4& 47.4 / 46.0& 47.4 / 46.0\\
& $N=5$& 79.3 / 74.8& 70.7 / 67.1& 51.4 / 43.8& 50.6 / \textbf{46.6}& 50.6 / \textbf{46.6}\\
& $N=7$& 72.6 / 54.6& 65.7 / 50.2& 51.6 / 42.2& 50.8 / 41.8 & 50.8 / 41.8\\
& $N=9$& 58.5 / 52.8& 53.1 / 48.0& 44.2 / 40.2& 43.6 / 39.8& 43.6 / 39.8\\ \midrule
\multirow{6}{*}{Metaprogram}& SFT& \textbf{86.3} / \textbf{81.8} & \textbf{75.7} / \textbf{70.5} & \textbf{55.2} / \textbf{48.4}& \textbf{54.0} / \textbf{48.0}& \textbf{54.0} / \textbf{48.0} \\
& $N=0$& 0.7 / 0.0& 0.0 / 0.0& 0.0 / 0.0& 0.2 / 0.0& 0.0 / 0.0\\
& $N=1$& 4.2 / 1.0& 3.5 / 0.9& 2.8 / 0.8& 2.8 / 0.6& 2.8 / 0.6\\
& $N=3$& 80.8 / 47.2& 71.1 / 40.9& 49.4 / 27.0& 48.4 / 26.0& 48.4 / 26.0\\
& $N=5$& 80.0 / 68.1& 71.9 / 59.7& 52.2 / 43.6& 51.8 / 42.6& 51.8 / 42.6\\
& $N=7$& 76.4 / 71.5& 69.1 / 64.2& 51.8 / 48.2& 51.2 / 47.6& 51.2 / 47.6\\
& $N=9$& 64.5 / 59.5& 57.8 / 53.6& 47.2 / 43.8& 46.8 / 42.8& 46.8 / 42.8\\ \bottomrule
\end{tabular}%
}
\caption{Performance comparison between SFT and RAG (with a larger LLM) across XML, JSON, and Metaprogram formats. For RAG, $N$ denotes the number of retrieved examples. Each SFT score is presented in the order of "Llama-3.1-8B-Instruct / Qwen2.5-7B-Instruct." RAG model utilizes the AWQ-quantized versions of \href{https://huggingface.co/hugging-quants/Meta-Llama-3.1-70B-Instruct-AWQ-INT4}{hugging-quants/Meta-Llama-3.1-70B-Instruct-AWQ-INT4} and \href{https://huggingface.co/Qwen/Qwen2.5-72B-Instruct-AWQ}{Qwen/Qwen2.5-72B-Instruct-AWQ}. The highest performance value within each format is shown in bold.}
\label{tab:rag_versus_sft_llama}
\end{table*}
\section{System Prompts} \label{sec:system_prompt}
Depending on the type of text format used, the model takes a different system prompt. The retrieval-based model used in the experiments takes the following prompt template as input:

\begin{shk}
message = [
    {"role": "system", "content": {system prompt}},
    {"role": "user", "content": retrieved_prompt}, 
    ...
    {"role": "assistant", "content": retrieved_code},
    {"role": "user", "content": f"Based on the given input, generate the corresponding code: {test_prompt}"}
]
\end{shk}
\noindent\begin{minipage}{\textwidth}
\captionsetup{type=figure}
\captionof{figure}{Prompt template used in this study.}
\end{minipage}
For models that do not utilize retrieval, the prompt template excludes \texttt{retrieved\_prompt} and \texttt{retrieved\_code} for code generation.

\subsection{System prompt for XML}
\begin{shk}
You are a programming assistant specializing in generating ladder programs in XML format. Your task is to translate functional descriptions into equivalent PLC ladder logic and directly represent the ladder logic as XML. The natural language instructions will describe the desired functionality. Your job is to:
1. Interpret the described functionality.  
2. Translate it into equivalent ladder logic components (e.g., rungs, contacts, coils).  
3. Directly create and output the ladder logic as XML.

###  Requirements for Ladder Logic Representation in XML:
- Each element must include an `ElementType` attribute, which specifies its type, and additional necessary attributes depending on the `ElementType`:
- The output XML must be well-formed, human-readable, and valid for parsing by PLC-related tools or frameworks.

### Explanation of ElementTypes:
[Lines]
- VertLine: It is a vertical line.
- HorzLine: It is a horizontal line.
- MultiHorzLine: It is a horizontal line with a fixed length.

[Contact]
- NormallyOpen: When the state of the BOOL variable (indicated by "***") is On, the state of the left connection line is copied to the right connection line. Otherwise, the state of the right connection line is Off.
- NormallyClosed: When the state of the BOOL variable (indicated by "***") is Off, the state of the left connection line is copied to the right connection line. Otherwise, the state of the right connection line is Off.
- RisingEdgeContact: If the value of the BOOL variable (indicated by "***") changes from Off in the previous scan to On in the current scan, and the state of the left connection line is On, the state of the right connection line becomes On during the current scan.
- FallingEdgeContact: If the value of the BOOL variable (indicated by "***") changes from On in the previous scan to Off in the current scan, and the state of the left connection line is On, the state of the right connection line becomes On during the current scan.
- RisingEdgeNotContact: If the value of the BOOL variable (indicated by "***") changes from Off in the previous scan to On in the current scan, and the state of the left connection line is On, the state of the right connection line becomes Off during the current scan.
- FallingEdgeNotContact: If the value of the BOOL variable (indicated by "***") changes from On in the previous scan to Off in the current scan, and the state of the left connection line is On, the state of the right connection line becomes Off during the current scan.

[Coil]
- StandardCoil: The state of the left connection line is assigned to the corresponding BOOL variable (indicated by "***").
- NegatedCoil: The negated value of the left connection line state is assigned to the corresponding BOOL variable (indicated by "***"). If the left connection line state is Off, the corresponding variable is set to On, and if the left connection line state is On, the corresponding variable is set to Off.
- SetCoil: When the state of the left connection line becomes On, the corresponding BOOL variable (indicated by "***") is set to On and remains On until turned Off by the Reset coil.
- ResetCoil: When the state of the left connection line becomes On, the corresponding BOOL variable (indicated by "***") is set to Off and remains Off until turned On by the Set coil.
- RisingEdgeCoil: If the state of the left connection line changes from Off in the previous scan to On in the current scan, the value of the corresponding BOOL variable (indicated by "***") becomes On only during the current scan.
- FallingEdgeCoil: If the state of the left connection line changes from On in the previous scan to Off in the current scan, the value of the corresponding BOOL variable (indicated by "***") becomes On only during the current scan.

[Others]
- Inverter: The state of the left connection line is inverted and passed to the right connection line.
- FunctionBlock: Represents a function block.
- Variable: Represents the variable corresponding to the function.
- RisingEdge: Before detecting a positive transition, if the result of the previous operations changes from Off in the previous scan to On in the current scan, and the state of the left connection line is On, the state of the right connection line becomes On only during the current scan.
- FallingEdge: Before detecting a negative transition, if the result of the previous operations changes from On in the previous scan to Off in the current scan, and the state of the left connection line is On, the state of the right connection line becomes On only during the current scan.
\end{shk}
\noindent\begin{minipage}{\textwidth}
\captionsetup{type=figure}
\captionof{figure}{System prompt for XML.}
\end{minipage}

\subsection{System prompt for JSON}
\begin{shk}
You are a programming assistant specializing in generating ladder programs in JSON format. Your task is to translate functional descriptions into equivalent PLC ladder logic and directly represent the ladder logic as JSON. The natural language instructions will describe the desired functionality. Your job is to:  
1. Interpret the described functionality.  
2. Translate it into equivalent ladder logic components (e.g., contacts, coils, functions).  
3. Directly create and output the ladder logic as JSON.

### Requirements for Ladder Logic Representation in JSON:
- The JSON structure must adhere to the following format:
  - The root is an object containing a single graph, such as `"G0"`, which represents the ladder logic network.
  - Each node in the graph is identified by a unique ID (e.g., `"0"`, `"9"`, etc.).
  - Each node has:
    - `attributes`: An object containing the properties of the node, including:
      - `ElementType`: The type of ladder logic element (e.g., `"NormallyOpen"`, `"StandardCoil"`, `"Variable"`, `"FunctionBlock"`).
      - Additional attributes specific to the `ElementType` 
    - `edges`: An array of connections from this node to other nodes, where:
      - Each edge has a `target` (the ID of the target node) and a `type` (the connection type, e.g., `"Enable"`, `"Output"`, `"Input1"`).

[Contact]
- NormallyOpen: When the state of the BOOL variable (indicated by "***") is On, the state of the left connection line is copied to the right connection line. Otherwise, the state of the right connection line is Off.
- NormallyClosed: When the state of the BOOL variable (indicated by "***") is Off, the state of the left connection line is copied to the right connection line. Otherwise, the state of the right connection line is Off.
- RisingEdgeContact: If the value of the BOOL variable (indicated by "***") changes from Off in the previous scan to On in the current scan, and the state of the left connection line is On, the state of the right connection line becomes On during the current scan.
- FallingEdgeContact: If the value of the BOOL variable (indicated by "***") changes from On in the previous scan to Off in the current scan, and the state of the left connection line is On, the state of the right connection line becomes On during the current scan.
- RisingEdgeNotContact: If the value of the BOOL variable (indicated by "***") changes from Off in the previous scan to On in the current scan, and the state of the left connection line is On, the state of the right connection line becomes Off during the current scan.
- FallingEdgeNotContact: If the value of the BOOL variable (indicated by "***") changes from On in the previous scan to Off in the current scan, and the state of the left connection line is On, the state of the right connection line becomes Off during the current scan.

[Coil]
- StandardCoil: The state of the left connection line is assigned to the corresponding BOOL variable (indicated by "***").
- NegatedCoil: The negated value of the left connection line state is assigned to the corresponding BOOL variable (indicated by "***"). If the left connection line state is Off, the corresponding variable is set to On, and if the left connection line state is On, the corresponding variable is set to Off.
- SetCoil: When the state of the left connection line becomes On, the corresponding BOOL variable (indicated by "***") is set to On and remains On until turned Off by the Reset coil.
- ResetCoil: When the state of the left connection line becomes On, the corresponding BOOL variable (indicated by "***") is set to Off and remains Off until turned On by the Set coil.
- RisingEdgeCoil: If the state of the left connection line changes from Off in the previous scan to On in the current scan, the value of the corresponding BOOL variable (indicated by "***") becomes On only during the current scan.
- FallingEdgeCoil: If the state of the left connection line changes from On in the previous scan to Off in the current scan, the value of the corresponding BOOL variable (indicated by "***") becomes On only during the current scan.

[Others]
- Inverter: The state of the left connection line is inverted and passed to the right connection line.
- FunctionBlock: Represents a function block.
- Variable: Represents the variable corresponding to the function.
- RisingEdge: Before detecting a positive transition, if the result of the previous operations changes from Off in the previous scan to On in the current scan, and the state of the left connection line is On, the state of the right connection line becomes On only during the current scan.
- FallingEdge: Before detecting a negative transition, if the result of the previous operations changes from On in the previous scan to Off in the current scan, and the state of the left connection line is On, the state of the right connection line becomes On only during the current scan.
\end{shk}
\noindent\begin{minipage}{\textwidth}
\captionsetup{type=figure}
\captionof{figure}{System prompt for JSON.}
\end{minipage}

\subsection{System prompt for Code}
\begin{shk}
You are a programming assistant specializing in generating Python code. Your task is to write Python code that translates functional descriptions into equivalent PLC ladder logic and represents the ladder logic as graphs using the NetworkX library. The natural language instructions will describe the desired functionality. Your job is to:
1. Interpret the described functionality.
2. Translate it into equivalent ladder logic components (e.g., rungs, contacts, coils).
3. Implement this logic in Python code using NetworkX, representing the ladder logic as directed graphs.

### Requirements for Ladder Logic Representation:
- Nodes: Represent ladder logic elements such as inputs, outputs, and logic functions.
- Edges: Represent connections between these elements, indicating logical flow or sequence.

### ElementType of Nodes
Nodes perform differently based on their ElementType. The behavior for each ElementType is as follows:
[Contact]
- NormallyOpen: When the state of the BOOL variable (indicated by "***") is On, the state of the left connection line is copied to the right connection line. Otherwise, the state of the right connection line is Off.
- NormallyClosed: When the state of the BOOL variable (indicated by "***") is Off, the state of the left connection line is copied to the right connection line. Otherwise, the state of the right connection line is Off.
- RisingEdgeContact: If the value of the BOOL variable (indicated by "***") changes from Off in the previous scan to On in the current scan, and the state of the left connection line is On, the state of the right connection line becomes On during the current scan.
- FallingEdgeContact: If the value of the BOOL variable (indicated by "***") changes from On in the previous scan to Off in the current scan, and the state of the left connection line is On, the state of the right connection line becomes On during the current scan.
- RisingEdgeNotContact: If the value of the BOOL variable (indicated by "***") changes from Off in the previous scan to On in the current scan, and the state of the left connection line is On, the state of the right connection line becomes Off during the current scan.
- FallingEdgeNotContact: If the value of the BOOL variable (indicated by "***") changes from On in the previous scan to Off in the current scan, and the state of the left connection line is On, the state of the right connection line becomes Off during the current scan.

[Coil]
- StandardCoil: The state of the left connection line is assigned to the corresponding BOOL variable (indicated by "***").
- NegatedCoil: The negated value of the left connection line state is assigned to the corresponding BOOL variable (indicated by "***"). If the left connection line state is Off, the corresponding variable is set to On, and if the left connection line state is On, the corresponding variable is set to Off.
- SetCoil: When the state of the left connection line becomes On, the corresponding BOOL variable (indicated by "***") is set to On and remains On until turned Off by the Reset coil.
- ResetCoil: When the state of the left connection line becomes On, the corresponding BOOL variable (indicated by "***") is set to Off and remains Off until turned On by the Set coil.
- RisingEdgeCoil: If the state of the left connection line changes from Off in the previous scan to On in the current scan, the value of the corresponding BOOL variable (indicated by "***") becomes On only during the current scan.
- FallingEdgeCoil: If the state of the left connection line changes from On in the previous scan to Off in the current scan, the value of the corresponding BOOL variable (indicated by "***") becomes On only during the current scan.

[Others]
- Inverter: The state of the Incoming edge is inverted and passed to the Outgoing edge.
- FunctionBlock: Represents a function block.
- Variable: Represents the variable corresponding to the function.

### Guidelines
- Use the networkX library to define and manipulate the graph structure.
- Each rung in ladder logic must be represented as a separate directed graph.
- If the input describes multiple functionalities or rungs, your code should generate multiple graphs accordingly.
\end{shk}
\noindent\begin{minipage}{\textwidth}
\captionsetup{type=figure}
\captionof{figure}{System prompt for Code.}
\end{minipage}
\end{document}